\documentclass{article}

\usepackage[preprint]{nips_2018}

\usepackage[utf8]{inputenc} 
\usepackage[T1]{fontenc}    
\usepackage{hyperref}       
\usepackage{url}            
\usepackage{booktabs}       
\usepackage{amsfonts}       
\usepackage{nicefrac}       
\usepackage{microtype}      
\usepackage{wrapfig}

\usepackage{times}
\usepackage{amsthm}
\usepackage{amsmath} 
\usepackage{amssymb}  
\usepackage{mathtools}
\usepackage{xcolor} 
\usepackage{ulem} 
\usepackage{multirow}
\usepackage{bbm}
\usepackage{comment}

\usepackage{color}
\usepackage{latexsym}
\usepackage{multicol}

\usepackage{algorithm}
\usepackage{algpseudocode}

\usepackage[font=small,labelfont=bf]{caption}
\usepackage{lipsum}
\usepackage{subcaption}



\newtheorem{definition}{Definition}
\newtheorem{problem}{Problem}

\usepackage[subnum]{cases}

\usepackage[backgroundcolor=White,
]{todonotes}

\newcommand{\Next}{\bigcirc}
\newcommand{\Always}{\Box}
\newcommand{\Event}{\diamondsuit}
\newcommand{\Implies}{\Rightarrow}
\newcommand{\Then}{\mathcal{T}}
\newcommand{\True}{\top}

\newcommand{\ignore}[1]{%
}

\usepackage{appendix}

\include{supplementary_material.tex}

\title{Automata-Guided Hierarchical Reinforcement Learning for Skill Composition}

\author{
  Xiao Li\thanks{} \qquad \qquad Yao Ma \qquad \qquad Calin Belta \\
  Department of Mechanical Engineering\\
  Boston University\\
  Boston, MA 02215 \\
  \texttt{{xli87,yaoma,cbelta}@bu.edu} \\
  \AND
}

\begin{document}

\maketitle

\begin{abstract}
Skills learned through (deep) reinforcement learning often generalizes poorly across domains and re-training is necessary when presented with a new task. We present a framework that combines techniques in \textit{formal methods} with \textit{reinforcement learning} (RL). The methods we provide allows for convenient specification of tasks with logical expressions, learns hierarchical policies (meta-controller and low-level controllers) with well-defined intrinsic rewards, and construct new skills from existing ones with little to no additional exploration. We evaluate the proposed methods in a simple grid world simulation as well as a more complicated kitchen environment in AI2Thor (\cite{ai2thor}).
\end{abstract}

\section{Introduction}
\label{sec:intro}



Hierarchical reinforcement learning (HRL) is an effective means of improving sample efficiency and achieving transfer among tasks. The goal is to obtain task-invariant low-level policies, and by re-training the meta-policy that schedules over the low-level policies, different skills can be obtain with less samples than training from scratch. \cite{heess2016learning} have adopted this idea in learning locomotor controllers and have shown successful transfer among simulated locomotion tasks. \cite{Oh2017} have utilized a deep hierarchical architecture for multi-task learning using natural language instructions.

Skill composition is the idea of constructing new skills out of existing ones (and hence their policies) with little to no additional learning. In stochastic optimal control, this idea has been adopted by \cite{todorov2009compositionality} and \cite{da2009linear} to construct provably optimal control laws based on linearly solvable Markov decision processes. \cite{haarnoja2018composable} have showed in simulated and real manipulation tasks that approximately optimal policies can result from adding the Q-functions of the existing policies.

Temporal logic(TL) is a formal language commonly used in software and digital circuit verification by \cite{baier2008principles} as well as formal synthesis by \cite{belta2017formal}. It allows for convenient expression of complex behaviors and causal relationships. TL has been used by \cite{tabuada2004linear}, \cite{fainekos2006translating}, \cite{fainekos2005hybrid} to synthesize provably correct control policies. \cite{aksaray2016q} have also combined TL with Q-learning to learn satisfiable policies in discrete state and action spaces. 

In this work, we focus on hierarchical skill learning and composition. Once a set of skills is acquired, we provide a technique that can synthesize new skills with little to no further interaction with the environment.  We adopt the syntactically co-safe truncated linear temporal logic(scTLTL) as the task specification language. Compared to most heuristic reward structures used in the RL literature, formal specification language has the advantage of semantic rigor and interpretability. Our main contributions are:

\begin{itemize}
\item Compared to existing skill composition methods, we are able to learn and compose logically complex tasks that would otherwise be difficult to analytically expressed as a reward function. We take advantage of the transformation between scTLTL formulas and finite state automata (FSA) to construct deterministic meta-controllers directly from the task specifications. We show that by adding one discrete dimension to the original state space, structurally simple parameterized policies such as feed-forward neural networks can be used to learn tasks that require complex temporal reasoning.

\item Intrinsic motivation has been shown to help RL agents learn complicated behaviors with less interactions with the environment (\cite{Singh2004}, \cite{Kulkarni2016}, \cite{jaderberg2016reinforcement}).  However, designing a well-behaved intrinsic reward that aligns with the extrinsic reward takes effort and experience. In our work, we construct intrinsic rewards directly from the input alphabets of the FSA, which guarantees that maximizing each intrinsic reward makes positive progress towards satisfying the entire task specification. From a user's perspective, the intrinsic rewards are constructed automatically from the TL formula without the need for further reward engineering. 

\item In our framework, each FSA represents a hierarchical policy with low-level controllers that can be re-modulated to achieve different tasks. Skill composition is accomplished by taking  the product of FSAs. Instead of interpolating/extrapolating among learned skills/latent features, our method is based on graph manipulation of the FSA. Therefore, the compositional outcome is much more transparent. At testing time, the behavior of the policy is strictly enforced by the FSA and therefore safety can be guaranteed if encoded in the specification. We introduce a method that allows learning of such hierarchical policies with any non-hierarchical RL algorithm. Compared with previous work on skill composition, we impose no constraints on the policy representation or the problem class.
\end{itemize}

\section{Preliminaries}
\label{sec:background}

\subsection{Reinforcement Learning}
\label{background_options}
We start with the definition of a Markov Decision Process.

\begin{definition}\label{def2}
An MDP is defined as a tuple $\mathcal{M} = \langle S,A,p(\cdot|\cdot,\cdot),r(\cdot,\cdot, \cdot)\rangle$, where $S\subseteq {\rm I\!R}^n$ is the state space ; $A \subseteq {\rm I\!R}^m$ is the action space ($S$ and $A$ can also be discrete sets); $p: S \times A \times S \to [0,1]$ is the transition function with $p(s^{\prime}|s,a)$ being the conditional probability density of taking action $a \in A$ at state $s \in S$ and ending up in state $s^{\prime} \in S$; $r: S \times A \times S \to {\rm I\!R}$ is the reward function with $r(s,a,s^\prime)$ being the reward obtained by executing action $a$ at state $s$ and transitioning to $s^\prime$. 
\end{definition}

Let $T$ be the horizon of the task. The optimal policy $\pi^\star: S \to A$ (or $\pi^\star: S \times A \to [0,1]$ for stochastic policies) that solves the MDP maximizes the expected return, i.e.

\begin{equation}\label{eq2B1}
\pi^\star = \underset{\pi}{\arg\max}\mathbb{E}^\pi[\sum_{t=0}^{T-1} r(s_t, a_t, s_{t+1})],
\end{equation}

\noindent where $\mathbb{E}^\pi[\cdot]$ is the expectation following $\pi$. The state-action value function is defined as 

\begin{equation}\label{eq2B2}
Q^\pi(s,a) = \mathbb{E}^\pi[\sum_{t=0}^{T-1} r(s_t, a_t, s_{t+1})|s_0=s, a_0=a]
\end{equation}
\noindent to be the expected return of choosing action $a$ at state $s$ and following $\pi$ onwards. Assuming the policy is greedy with respect to $Q$ i.e. $\pi(s) = \underset{a}{\arg\max}Q(s,a)$, then at convergence, Equation~\eqref{eq2B2} yields

\begin{equation}\label{eq2B3}
Q^\star(s_t,a_t) = \mathbb{E}^{\pi_{explore}}[r(s_t, a_t, s_{t+1}) + \gamma \underset{a}{\max}Q^\star(s_{t+1}, a)]
\end{equation}

\noindent where $Q^\star$ is the optimal state-action value function, $\gamma \leq 1$ is a discount factor that favors near term over long term rewards if smaller than 1. $\pi_{explore}$ can be any exploration policy (will sometimes be omitted for simplicity of presentation). This is the setting that we will adopt for the remainder of this work.


\subsection{scTLTL and Finite State Automata}
\label{sec:background_fsa}

We consider tasks specified with \textit{syntactically co-safe Truncated Linear Temporal Logic} (scTLTL) which is a fragment of truncated linear temporal logic(TLTL) (\cite{li2016reinforcement}). The set of allowed operators are 

\begin{equation}\label{eq2A1}
\begin{split}
\phi := \ & \True \,\,|\,\, f(s) < c \,\,| \,\, \neg \phi \,\,|\,\, \phi \wedge \psi \,\,|\,\, \\ & \Event \phi  \,\,|\,\, \phi \, \mathcal{U} \, \psi \,\,|\,\, \phi\, \Then\, \psi \,\,|\,\, \Next \phi \,\, 
\end{split}
\end{equation}
where $\True$ is the True Boolean constant. $f(s) < c$ is a predicate.
$\neg$~(negation/not), $\wedge$~(conjunction/and) are
Boolean connectives.
$\Event$~(eventually),  $\mathcal{U}$~(until), $\Then$~(then), $\Next$~(next),
are temporal operators.$\Implies$ (implication) and and $\vee$~(disjunction/or) can be derived from the above operators. Compared to TLTL, we excluded the $\Always$~(always) operator to maintain a one to one correspondence between an scTLTL formula and a finite state automaton (FSA) defined below.


\begin{definition}\label{def1} 
An FSA\footnote{Here we slightly modify the conventional definition of FSA and incorporate the probabilities in Equations~\eqref{eq2A2}. For simplicity, we continue to adopt the term FSA.} is defined as a tuple  $\mathcal{A}=\langle \mathbb{Q}, \Psi, q_0, p(\cdot | \cdot), \mathcal{F} \rangle $, where $\mathbb{Q}$ is a set of automaton states; $\Psi$ is the input alphabet; $q_0 \in \mathbb{Q}$ is the initial state; $p:\mathbb{Q} \times \mathbb{Q} \rightarrow [0,1]$ is a conditional probability defined as 

\begin{equation}\label{eq2A2}
p_\phi(q_j | q_i) = \begin{cases}
1 & \psi_{q_i, q_j} \textrm{ is true} \\
 0 & \text{otherwise}.
 \end{cases} 
 \quad or \quad 
p_\phi(q_j | q_i, s) = \begin{cases}
1 & \rho(s,\psi_{q_i, q_j})>0\\
 0 & \text{otherwise}.
 \end{cases} 
\end{equation}

\noindent $\mathcal{F}$ is a set of final automaton states.

\end{definition}

We denote $\psi_{q_i, q_j} \in \Psi$ the predicate guarding the transition from $q_i$ to $q_j$. Because $\psi_{q_i, q_j}$ is a predicate without temporal operators, the robustness $\rho(s_{t:t+k}, \psi_{q_i, q_j})$ is only evaluated at $s_t$. Therefore, we use the shorthand $\rho(s_{t}, \psi_{q_i, q_j}) = \rho(s_{t:t+k}, \psi_{q_i, q_j})$. We abuse the notation $p()$ to represent both kinds of transitions when the context is clear. For each scTLTL formula, one can construct a corresponding FSA $\mathcal{A}_\phi$. An example of an FSA is provided in Section C.1 in the supplementary material. The translation from TLTL formula to FSA to can be done automatically with available packages like Lomap (\cite{lomap}). 

There exists a real-valued function $\rho(s_{0:T}, \phi)$ called robustness degree that measures the level of satisfaction of trajectory $s_{0:T}$ (here $s_{0:T}=[s_0, ..., s_T]$ is the state trajectory from time 0 to $T$) with respect to a scTLTL formula $\phi$. $\rho(s_{0:T}, \phi)>0$ indicates that $s_{0:T}$ satisfies $\phi$ and vice versa (full semantics of scTLTL are provided in Section A in supplementary material).


\section{Problem Formulation}
\label{sec:problem}

\begin{problem}\label{p1}
Given an MDP in Definition~\ref{def2} with unknown transition dynamics $p(s^\prime|s,a)$ and a scTLTL specification $\phi$ as in Definition~\ref{def1}, find a policy $\pi^\star_\phi$ such that


\begin{equation}\label{eq3A1}
\pi^\star_\phi = \underset{\pi_\phi}{\arg \max}\mathbb{E}^{\pi_\phi}[\mathbbm{1}(\rho(s_{0:T}, \phi)>0)]. 
\end{equation}
\noindent where $\mathbbm{1}(\rho(s_{0:T}, \phi)>0)$  is an indicator function with value $1$ if $\rho(s_{0:T}, \phi)>0$ and $0$ otherwise. $\pi^\star_\phi$ is said to satisfy $\phi$.
\end{problem}

\noindent Problem~\ref{p1} defines a policy search problem where the trajectories resulting from following the optimal policy should satisfy the given scTLTL formula in expectation. It should be noted that there typically will be more than one policy that satisfies Equation~\eqref{eq3A1}. We use a discount factor to reduce the number of satisfying policies to one (one that yields a satisfying trajectory in the least number of steps). Details will be discussed in the next section.

\begin{problem} \label{p2}
Given two scTLTL formula $\phi_1$ and $\phi_2$ along with policy $\pi_{\phi_1}$ that satisfies $\phi_1$ and $\pi_{\phi_2}$ that satisfies $\phi_2$ (and their corresponding state-action value function $Q(s,q_1,a)$ and $Q(s,q_2,a$)), obtain a policy $\pi_{\phi}$ that satisfies $\phi=\phi_1 \wedge \phi_2$. 
\end{problem}

\noindent Problem~\ref{p2} defines the problem of skill composition. Given two policies each satisfying a scTLTL specification, construct the policy that satisfies the conjunction of the given specifications. Solving this problem is useful when we want to break a complex task into simple and manageable components, learn a policy that satisfies each component and "stitch" all the components together so that the original task is satisfied. It can also be the case that as the scope of the task grows with time, the original task specification is amended with new items. Instead of having to re-learn the task from scratch, we can learn only policies that satisfies the new items and combine them with the old policy. 


\section{FSA Augmented MDP}
\label{augmentedMDP}

Problem~\ref{p1} can be solved with any RL algorithm using robustness as the terminal reward as is done by \cite{li2016reinforcement}. However, doing so the agent suffers from sparse feedback because a reward signal can only be obtained at the end of each episode. To address this problem as well as setting up ground for automata guided HRL, we introduce the FSA augmented MDP

\begin{definition}\label{def3}
 An FSA augmented MDP corresponding to scTLTL formula $\phi$ (constructed from FSA $\langle \mathbb{Q}_\phi, \Psi_\phi, q_{\phi,0}, p_\phi(\cdot | \cdot), \mathcal{F}_\phi \rangle$ and MDP $\langle S,A,p(\cdot|\cdot,\cdot),r(\cdot,\cdot, \cdot)\rangle$) is defined as $\mathcal{M} _\phi= \langle \tilde{S}, A, \tilde{p}(\cdot|\cdot,\cdot),\tilde{r}(\cdot, \cdot),  \mathcal{F}_\phi \rangle$ where $\tilde{S} \subseteq S \times \mathbb{Q}_{\phi}$, $\tilde{p}(\tilde{s}'|\tilde{s},a)$ is the probability of transitioning to $\tilde{s}^\prime$ given $\tilde{s}$ and $a$,
 \begin{equation}\label{eq3A2}
 \tilde{p}(\tilde{s}'|\tilde{s},a) = p\big((s', q')|(s,q), a\big)
 = \begin{cases}
 p(s'|s,a) & p_\phi(q'|q,s) =1 \\
 0 & \text{otherwise}.
 \end{cases}
 \end{equation}

\noindent $p_\phi$ is defined in Equation~\eqref{eq2A2}. $\tilde{r}: \tilde{S} \times \tilde{S} \to {\rm I\!R}$ is the FSA augmented reward function, defined by 


\begin{equation}\label{eq3A3}
\tilde{r}(\tilde{s},\tilde{s}') =\mathbbm{1}\big( \rho(s',D_{\phi}^{q})>0\big), 
\end{equation}
\noindent where $D_{\phi}^{q}=\bigvee_{q^\prime \in\Omega_{q}} \psi_{q, q^\prime}$ represents the disjunction of all predicates guarding the transitions that originate from $q$ ($\Omega_{q}$ is the set of automata states that are connected with $q$ through outgoing edges). 
\end{definition}

The goal is to find the optimal policy that maximizes the expected sum of discounted return, i.e.

\begin{equation}\label{eq3A4}
 \pi^\star = \underset{\pi}{\arg\max} \mathbb{E}^{\pi}\left[\sum_{t=0}^{T-1} \gamma^{t+1}\tilde{r}(\tilde{s}_t,\tilde{s}_{t+1})\right], 
\end{equation}

\noindent where $\gamma < 1$ is the discount factor, $T$ is the time horizon.


The reward function in Equation~\eqref{eq3A3} encourages the system to exit the current automata state and move on to the next, and by doing so eventually reach the final state $q_f$ (property of FSA) which satisfies the TL specification and hence Equation~\eqref{eq3A1}. The discount factor in Equation~\eqref{eq3A4} reduces the number of satisfying policies to one. 



The FSA augmented MDP can be constructed with any standard MDP and a scTLTL formula, and can be solved with any off-the-shelf RL algorithm. By directly learning the flat policy $\pi$ we bypass the need to define and learn each sub-policy separately. After obtaining the optimal policy $\pi^\star$, the optimal sub-policy for any $q_i$ can be extracted by executing $\pi^\star(s_t,q_i)$ without transitioning the automata state, i.e. keeping $q_i$ fixed. The sub-policy is thus

\begin{equation}\label{eq3A5}
\pi^\star(s_t,q_i) = \underset{a_t}{\arg\max}Q^\star(s_t, q_i, a_t), 
\end{equation}

\noindent where

\begin{equation}\label{eq3A6}
Q^\star(s_t, q_i, a_t) = \mathbb{E}^{\pi^\star}[\sum_{t=0}^{T-1}\mathbbm{1}\big(\rho(s_{t+1},D_{\phi}^{q_{i}})>0\big)]
= \mathbb{E}\left[\mathbbm{1}\big(\rho(s_{t+1},D_{\phi}^{q_{i}})>0\big) + \gamma \underset{a}{\max}Q^\star(s_{t+1}, q_i, a)\right]. 
\end{equation}


\section{Automata Guided Skill Composition}
\label{spec_amend}

In section, we provide a solution for Problem~\ref{p2} by constructing the FSA of $\phi$ from that of $\phi_1$ and $\phi_2$ and using $\phi$ to synthesize the policy for the combined skill. We start with the following definition. 

\begin{definition}\label{def4}\footnote{details can be found in~\cite{product}}
Given $\mathcal{A}_{\phi_1}=\langle \mathbb{Q}_{\phi_1}, \Psi_{\phi_1}, q_{1,0}, p_{\phi_1}, \mathcal{F}_{\phi_1} \rangle$ and  $\mathcal{A}_{\phi_2}=\langle \mathbb{Q}_{\phi_2}, \Psi_{\phi_2}, q_{2,0}, p_{\phi_2}, \mathcal{F}_{\phi_2} \rangle$ corresponding to formulas $\phi_1$ and $\phi_2$,  the FSA of $\phi=\phi_1 \wedge \phi_2$ is the \textit{product automaton} of $\mathcal{A}_{\phi_1}$ and $\mathcal{A}_{\phi_1}$, i.e.  $\mathcal{A}_{\phi=\phi_1 \wedge \phi_2} = \mathcal{A}_{\phi_1} \times \mathcal{A}_{\phi_2} =  \langle \mathbb{Q}_\phi, \Psi_\phi, q_0, p_\phi, \mathcal{F}_\phi \rangle$ where $\mathbb{Q}_{\phi} \subseteq \mathbb{Q}_{\phi_1} \times \mathbb{Q}_{\phi_2}$ is the set of product automaton states, $q_0 = (q_{1,0}, q_{2,0})$ is the product initial state, $\mathcal{F} \subseteq \mathcal{F}_{\phi_1} \cap \mathcal{F}_{\phi_2}$ are the final accepting states. Following Definition~\ref{def1}, for states $q=(q_1, q_2) \in \mathbb{Q}_\phi$ and $q^\prime = (q_1^\prime, q_2^\prime) \in \mathbb{Q}_\phi$, the transition probability $p_\phi$ is defined as 

\begin{equation}\label{eqF1}
p_\phi(q^\prime| q) = \begin{cases}
1 & p_{\phi_1}(q_1^\prime|q_1)p_{\phi_2}(q_2^\prime|q_2)=1 \\
 0 & \text{otherwise}.
 \end{cases}
\end{equation}

\end{definition}

\noindent An example of product automaton is provided in Section C.2 in the supplementary material.


For $q=(q_1, q_2) \in \mathbb{Q}_\phi$, let $\Psi_q$,  $\Psi_{q_1}$ and $\Psi_{q_2}$denote the set of predicates guarding the edges originating from $q$, $q_1$ and $q_2$ respectively. Equation~\eqref{eqF1} entails that a transition at $q$ in the product automaton $\mathcal{A}_\phi$ exists only if corresponding transitions at $q_1$, $q_2$ exist in $\mathcal{A}_{\phi_1}$and $\mathcal{A}_{\phi_2}$ respectively. Therefore, $\psi_{q,q^\prime} = \psi_{q_1, q_1^\prime} \wedge \psi_{q_2, q_2^\prime}$, for  $\psi_{q,q^\prime}\in\Psi_q, \psi_{q_1, q_1^\prime}\in\Psi_{q_1}, \psi_{q_2, q_2^\prime}\in\Psi_{q_2}$ (here $q_i^\prime$ is a state such that $p_{\phi_i}(q_i^\prime | q_i) = 1$).  Following Equation~\eqref{eq3A6}, 

\begin{equation}\label{eq3C3}
\begin{split}
&Q^\pi(s_t, q_t, a_t) = \mathbb{E}^\pi\big [ \sum^{T-1}_{t=0} \gamma^{t+1}\mathbbm{1}\big (\rho(s_{t+1}, D^{q_t}_\phi)>0\big)  \big], \textrm{where }  D^{q_t}_\phi = \bigvee_{q_{1,t}^\prime, q_{2,t}^\prime} ( \psi_{q_{1,t}, q_{1,t}^\prime} \wedge \psi_{q_{2,t}, q_{2,t}^\prime}) \\ 
& \textrm{ and } q_{1,t}^\prime, q_{2,t}^\prime \textrm{ don't equal to } q_{1,t}, q_{2,t} \textrm{ at the same time (to avoid self looping edges)}.
\end{split}
\end{equation}

\noindent Here $q_{i,t}$ is the FSA state of $\mathcal{A}_{\phi_i}$ at time $t$. $q_{i,t}^\prime \in \Omega_{q_{i,t}}$ are FSA states that are connected to $q_{i,t}$ through an outgoing edge. It can be shown that 

\begin{equation}\label{eq3C8}
Q^\pi(s_t, q_t, a_t) = Q^\pi_{q_{1,t}}(s_t, q_t, a_t) + Q^\pi_{q_{2,t}}(s_t, q_t, a_t) - Q^\pi_{q_{1,t} \wedge q_{2,t}}(s_t, q_t, a_t)
\end{equation}

\noindent where

\begin{equation}\label{eq3C9}
Q^\pi_{q_{i,t}}(s_t, q_t, a_t) = \mathbb{E}^\pi\left[\sum_{t=0}^{T-1} \gamma^{t+1}\mathbbm{1}\big(\rho(s_{t+1},D_{\phi_i}^{q_{i,t}})>0\big)\right] \textrm{, } i=1,2
\end{equation}

\begin{equation}\label{eq3C10}
Q^\pi_{q_{1,t} \wedge q_{2,t}}(s_t, q_t, a_t)=\mathbb{E}^\pi\big [ \sum^{T-1}_{t=0}\gamma^{t+1} \mathbbm{1}(\rho(s_{t+1},D_{\phi_1}^{q_{1,t}})>0)\mathbbm{1}(\rho(s_{t+1},D_{\phi_2}^{q_{2,t}})>0)\big)\big].
\end{equation}

\noindent We provide the derivation in Section B in the supplementary material.

Equation~\eqref{eq3C9} takes similar form as Equation~\eqref{eq3A6}. Since we have already learned $Q^\star(s_t,q_{1,t},a_t)$ and $Q^\star(s_t,q_{2,t},a_t)$, and $Q_{q_{1,t} \wedge q_{2,t}}(s_t, q_t, a_t)$ is nonzero only when there are states $s_t$ where $D^{q_{1,t}}_{\phi_1} \wedge D^{q_{2,t}}_{\phi_2}$ is true, we should obtain a good initialization of $Q^\star(s_t,q_t,a_t)$ by adding $Q^\star(s_t,q_{1,t},a)$ and $Q^\star(s_t,q_{2,t},a_t)$ (similar technique is adopted by \cite{haarnoja2018composable}). This addition of local $Q$ functions is in fact an optimistic estimation of the global $Q$ function, the properties of such Q-decomposition methods are studied by \cite{russell2003q}. 

Here we propose an algorithm to obtain the optimal composed Q function $Q^\star(s_t,q_t,a_t)$ given the already learned $Q^\star(s_t,q_{1,t},a_t)$, $Q^\star(s_t,q_{2,t},a_t)$ and the data collected while training them.

\begin{algorithm}
\caption{FSA guided skill composition}
\label{alg:2}
\begin{algorithmic}[1]
\State \textbf{Inputs}: The learned Q functions $Q^\star(s_t,q_{1,t},a_t)$ and $Q^\star(s_t,q_{2,t},a_t)$, replay pool $\mathcal{B}$ collected when training $Q^\star_i, i=1,2$. The product FSA $\mathcal{A}_\phi$
\State Initialize  $Q(s_t, q_t, a_t) \leftarrow Q^\star(s_t,q_{1,t},a_t) + Q^\star(s_t,q_{2,t},a_t)$
\State Initialize $Q_{q_{1,t} \wedge q_{2,}}(s_t, q_t, a_t)$
\State $Q_{q_{1,t} \wedge q_{2,}} \leftarrow update(Q_{q_{1,t} \wedge q_{2,}},\mathcal{A}_\phi, \mathcal{B}, r_{q_{1,t}\wedge q_{2,}})$ \Comment $r_{q_{1,t} \wedge q_{2,t}}=\mathbbm{1}(\rho(s_{t+1},D_{\phi_1}^{q_{1,t}})>0)\mathbbm{1}(\rho(s_{t+1},D_{\phi_2}^{q_{2,t}})>0) \textrm{ as in Equation~\eqref{eq3C10}}$
\State $Q(s_t, q_t, a_t) \leftarrow Q^\star(s_t,q_{1,t},a_t) + Q^\star(s_t,q_{2,t},a_t) - \tilde{Q}_{q_{1,t} \wedge q_{2,t}}$
\State $Q(s_t, q_t, a_t) \leftarrow update(Q(s_t, q_t, a_t),\mathcal{A}_\phi, \mathcal{B}, r_{q_t})$ \Comment $r_{q_t}=\mathbbm{1}(\rho(s_{t+1},D_{\phi_1}^{q_{1,t}})>0)+\mathbbm{1}(\rho(s_{t+1},D_{\phi_2}^{q_{2,t}})>0) - \mathbbm{1}(\rho(s_{t+1},D_{\phi_1}^{q_{1,t}})>0)\mathbbm{1}(\rho(s_{t+1},D_{\phi_2}^{q_{2,t}})>0) \textrm{ as in Equation~\eqref{eq3C8} - \eqref{eq3C10}}$\\
\Return $Q(s_t, q, a_t)$
\end{algorithmic}
\end{algorithm}

\noindent The Q functions in Algorithm~\ref{alg:2} can be grid representation or a parametrized function. The $update(\cdot, \cdot, \cdot, \cdot)$  function that takes in a Q-function, the product FSA, stored replay buffer and a reward, and performs off-policy Q update. If the initial state distribution remains unchanged, Algorithm~\ref{alg:2} should provide a decent estimate of the composed Q function without needing to further interact with the environment.The intuition is that the experience collected from training $Q^\star(s_t,q_{1,t},a_t)$ and $Q^\star(s_t,q_{2,t},a_t)$ should have well explored the regions in state space that satisfy $\phi_1$ and $\phi_2$, and hence also explored the regions that satisfy $\phi_1 \wedge \phi_2$. Having obtained $Q(s_t,q_t,a_t)$, a greedy policy can be extracted in similar ways to DQN (\cite{Mnih2015}) for discrete actions or DDPG (\cite{Silver2014}) for continuous actions. Details of Algorith~\ref{alg:2} are provided in Section D.5 in the supplementary materal.
\section{Case Studies}
\label{sec:case_studies}

We evaluate the proposed methods in two types of environments. The first is a grid world environment that aims to illustrate the inner workings of our method. The second is a kitchen environment simulated in AI2Thor (\cite{ai2thor}). 

\subsection{Grid World}
\label{subsec:grid_world}

\begin{figure*}
\vspace{-0.1in}
\begin{center}
\includegraphics[width=1.\linewidth]{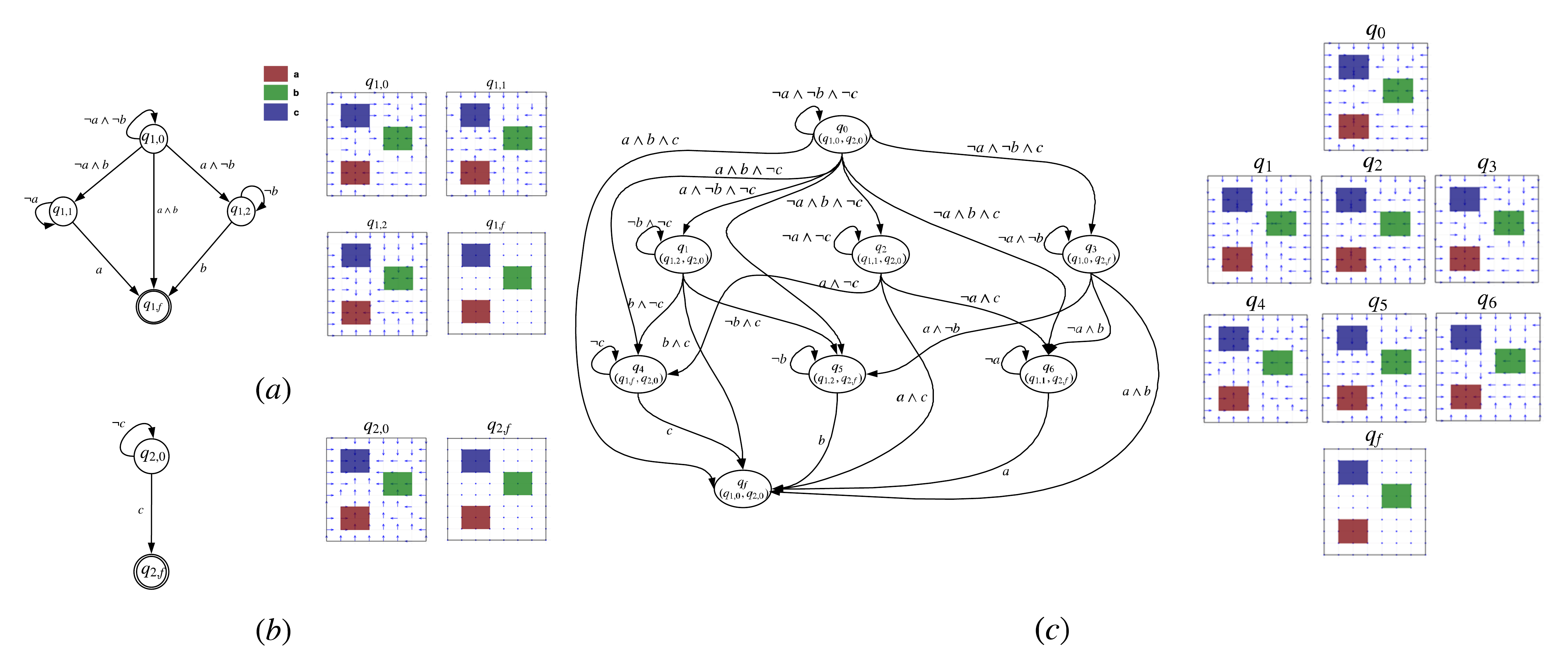}
\vspace{-0.1in}
\caption{FSA and policy for \textbf{(a)} $\phi_1=\diamondsuit a \wedge \diamondsuit b$. \textbf{(b)} $\phi_2=\diamondsuit c$. \textbf{(c)} $\phi = \phi_1 \wedge \phi_2$. The agent moves in a $8 \times 10$ gridworld with 3 labeled regions. The agent has actions [\textit{up}, \textit{down}, \textit{left}, \textit{right}, \textit{stay}] where the directional actions are represented by arrows, \textit{stay} is represented by the blue dot.}\label{fig:2}
\end{center}
\end{figure*}

Consider an agent that navigates in a $8 \times 10$ grid world. Its MDP state space is $S: X \times Y $ where $x,y \in X, Y$ are its integer coordinates on the grid. The action space is $A:$ [\textit{up}, \textit{down}, \textit{left}, \textit{right}, \textit{stay}]. The transition is such that for each action command, the agent follows that command with probability 0.8 or chooses a random action with probability 0.2. We train the agent on two tasks, $\phi_1=\diamondsuit a \wedge \diamondsuit b$ and $\phi_2=\diamondsuit c$. In English, $\phi_1$ expresses the requirement that for the horizon of task, regions $a$ and $b$ need to be reached at least once. The regions are defined by the predicates $a=(1<x<3) \wedge (1<y<3)$ and $b=(4<x<6) \wedge (4<y<6)$. Because the coordinates are integers, $a$ and $b$ define a point goal rather than regions. $\phi_2$ expresses a similar task for $c=(1<x<3) \wedge (6<y<8)$. Figure~\ref{fig:2} shows the FSA for each task. 

We apply standard tabular Q-learning (\cite{Watkins}) on the FSA augmented MDP of this environment. For all experiments, we use a discount factor of 0.95, learning rate of 0.1, episode horizon of 200 steps, a random exploration policy and a total number of 2000 update steps which is enough to reach convergence.

Figure~\ref{fig:2} (a) and (b) show the learned optimal policies extracted by $\pi^\star(x,y,q) = \underset{a}{\arg\max}Q^\star(x,y,q,a)$. We plot $\pi^\star(x,y,q_i)$ for each $q_i$ and observe that each represents a sub-policy whose goal is given by Equation~\eqref{eq3A3}. The FSA effectively acts as a meta-policy. We are able to obtain such meaningful hierarchy without having to explicitly incorporate it in the learning process.

Figure~\ref{fig:2} (c) shows the composed FSA and policy using Algorithm~\ref{alg:2}. Prior to composition, we normalized the Q functions by dividing each by its max value put them in the same range. This is possible because the Q values of both policies have the same meaning (expected discounted edge distance to $q_f$ on the fSA).In this case the initialization step (step 2) is sufficient to obtain the optimal composed policy without further updating necessary. The reason is that there are no overlaps between regions $a,\, b,\, c$, therefore $r_{q_1 \wedge q_2}=0$ for all states and actions which renders steps 3, 4, 5 unnecessary. We found that step 6 in Algorithm~\ref{alg:2} is also not necessary here.


\subsection{AI2Thor}
\label{subsec:ai2thor}


In this section, we apply the proposed methods in a simulated kitchen environment. The goal is to find a user defined object (e.g. an apple) and place it in a user defined receptacle (e.g. the fridge). Our main focus for this experiment is to learn a high level decision-making policy and therefore we assume that the agent can navigate to any desired location. 

There are a total of 17 pickupable objects and 39 receptacle objects which we index from 0 to 55. Our state space depends on these objects and their properties/states. We have a set of 62 discrete actions \{\textit{pick}, \textit{put}, \textit{open}, \textit{close}, \textit{look up}, \textit{look down}, \textit{navigate(id)}\} where \textit{id} can take values from 0 to 55. Detailed descriptions of the environment, state and action spaces are provided in Sections D.1 , D.2 and D.3 of the supplementary material.

We start with a relatively easy task of "\textit{find and pick up the apple and put it in the fridge}"(which we refer to as task 1) and extend it to "\textit{find and pick up any user defined object and put it in any user defined receptacle}" (which we refer to as task 2). For each task, we learn with three specifications with increasing prior knowledge encoded in the scTLTL formula. The specifications are referred to as $\phi_{i,j}$ with $i \in \{1,2\}$ denoting the task number and $j \in \{1,2,3\}$ denoting the specification number. The higher the $j$ more prior knowledge is encoded. We also explore the combination of the intrinsic reward defined in the FSA augmented MDP with a heuristic penalty. Here we penalize the agent for each failed action and denote the learning trials with penalty by $\phi_{i,j}^\star$.  To evaluate automata guided skill composition, we combine task 1 and task 2 and require the composed policy to accomplish both tasks during an episode (we refer to this task as composition task). Details on the specifications are provided in Section D.4 of the supplementary material.

We use a feed forward neural network as the policy and DDQN (\cite{van2016deep}) with prioritized experience replay (\cite{schaul2015prioritized}) as the learning algorithm. We found that adaptively normalizing the Q function with methods proposed in (\cite{van2016learning}) helps accelerate task composition. Algorithm details are provided in Section D.5 of the supplementary material. For each task, we evaluate the learned policy at various points along the training process by running the policy without exploration for 50 episodes with random initialization. Performance is evaluated by the average task success rate and episode length (if the agent can quickly accomplish the task). We also include the action success rate (if the agent learns not to execute actions that will fail) during training as a performance metric. 

\begin{figure*}[h] 
\vspace{-0.1in}
\begin{center}
\includegraphics[width=0.8\linewidth]{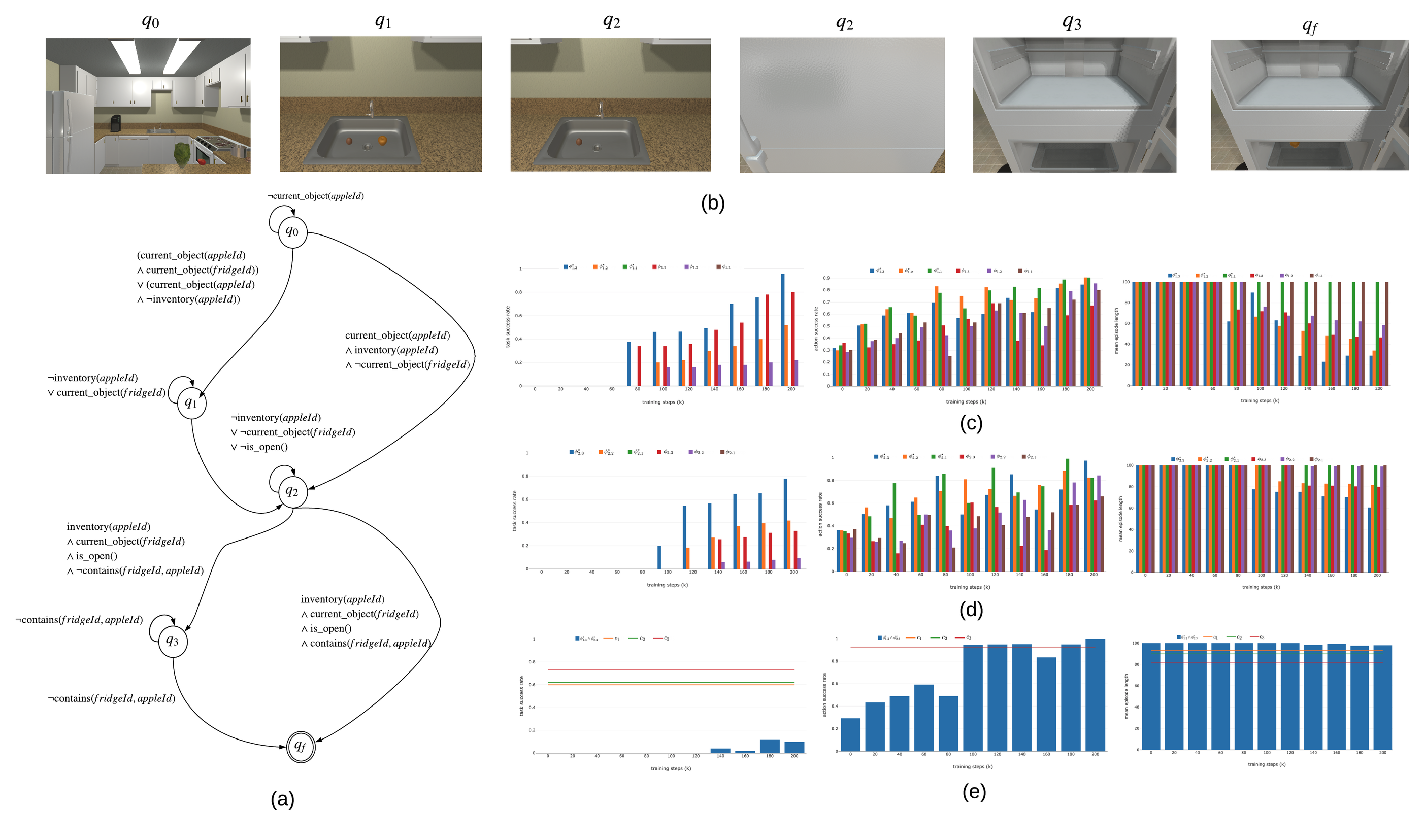}
\vspace{-0.1in}
\caption{\textbf{(a)} FSA for specification $\phi_{1.3}$. \textbf{(b)} Agent's first person view of the environment at each transition to a new FSA state (the apple in the last image is placed on the first bottom shelf). Task success rate, action success rate and mean episode length for (c) task 1. (d) task 2. (e) composition task}\label{fig:3}
\end{center}
\end{figure*}

Figure~\ref{fig:3}(a) shows the FSA of specification $\phi_{1.3}$, and Figure~\ref{fig:3}(b) illustrates the agent's first person view at states where transition on the FSA occurs. Note that navigating from the sink (with apple picked up) to the fridge does not invoke progress on the FSA because such instruction is not encoded in the specification. Figure~\ref{fig:3}(c) shows the learning performance of task 1. We can see that the more elaborate the specification, the higher the task success rate which is as expected ($\phi_{1.1}$ and $\phi^\star_{1.1}$ fail to learn the task due to sparse reward). It can also be observed that the action penalty helps facilitate the agent to avoid issuing failing actions and in turn reduces the steps necessary to complete the task.

Figure~\ref{fig:3}(d) shows the results for task 2. Most of the conclusions from task 1 persists. The success rate for task 2 is lower due to the added complexity of the task. The mean episode length is significantly larger than task 1. This is because the object the agent is required to find is often initialized inside receptacles, therefore the agent needs to first find the object and then proceed to completing the task. This process is not encoded in the specification and hence rely solely on exploration. An important observation here is that learning with action penalty significantly improves the task success rate. The reason is also that completing task 2 may requires a large number of steps when the object is hidden in receptacles, the agent will not have enough time if the action failure rate is high.  

Figure~\ref{fig:3}(e) shows the performance of automata guided skill composition. Here we present results of progressively running Algorithm~\ref{alg:2}. In the figure, $c_1$ represents running only the initialization step (step 2 in the algorithm), $c_2$ represents running the initialization and compensation steps (steps 3, 4, 5) and $c_3$ is running the entire algorithm. As comparison, we also learn this task from scratch with FSA augmented MDP with the specification $\phi^\star_{1.3} \wedge \phi^\star_{2.3}$. From the figures we can see that the action success rate is not effected by task complexity. Overall, the composed policy considerably outperforms the trained policy (the resultant product FSA for this task has 23 nodes and 110 edges, therefore is expected to take longer to train). Simply running the initiation step $c_1$ already results in a decent policy. Incorporating the compensation step in $c_2$ did not provide a significant improvement. This is most likely due to the lack of MDP states $s$ where $\mathbbm{1}(\rho(s,D_{\phi_1}^{q_{1}})>0)\mathbbm{1}(\rho(s,D_{\phi_2}^{q_{2}})>0)\big) \neq 0$ ($q_1 \in \mathbb{Q}_{\phi^\star_{1.3}}$, $q_2 \in \mathbb{Q}_{\phi^\star_{2.3}}$). However, $c_3$ improves the composed policy by a significant margin because this step fine tunes the policy with the true objective and stored experiences.  We provide additional discussions in Section D.6 of the supplementary material.


\section{Conclusion}
\label{sec:conclusion}

We present a framework that integrates the flexibility of reinforcement learning with the explanability and semantic rigor of formal methods. In particular, we allow task specification in scTLTL - an expressive formal language, and construct a product MDP that possesses an intrinsic hierarchical structure. We showed that applying RL methods on the product MDP results in a hierarchical policy whose sub-policies can be easily extracted and re-combined to achieve new tasks in a transparent fashion. In practice, the authors have particularly benefited from the FSA in terms of specification design and behavior prediction in that mistakes in task expression can be identified before putting in the time and resources for training.

\bibliographystyle{plainnat}
\bibliography{references}

\end{document}